\documentclass[sigconf,natbib=true,anonymous=false]{acmart}
\renewcommand\footnotetextcopyrightpermission[1]{} 
\usepackage{multibib}
\usepackage{graphicx}
\usepackage{tabularx}
\usepackage{supertabular,booktabs}
\usepackage{tikz}
\usepackage{pgfplots}
\usepackage{hyperref}
\usepackage{algorithm}
\usepackage{algpseudocode}
\usepackage{subcaption}
\usetikzlibrary{shapes.geometric, arrows}
\AtBeginDocument{%
  \providecommand\BibTeX{{%
    \normalfont B\kern-0.5em{\scshape i\kern-0.25em b}\kern-0.8em\TeX}}}

\begin{document}

\title{Text Representation Enrichment Utilizing Graph based Approaches: Stock Market Technical Analysis Case Study}


\author{Sara Salamat}
\authornote{Both authors contributed equally to this research.}
\author{Nima Tavassoli}
\authornotemark[1]
\author{Behnam Sabeti}
\author{Reza Fahmi}
\email{{sara, nima, behnam, reza}@eveince.com}
\affiliation{%
  \institution{Eveince}
  \streetaddress{WALLSTRAßE 18, 10179}
  \city{Berlin}
  \country{Germany}
}

\renewcommand{\shortauthors}{Salamat and Tavassoli, et al.}

\begin{abstract}
Graph neural networks (GNNs) have been utilized for various natural language processing (NLP) tasks lately. The ability to encode corpus-wide features in graph representation made GNN models popular in various tasks such as document classification. One major shortcoming of such models is that they mainly work on homogeneous graphs, while representing text datasets as graphs requires several node types which leads to a heterogeneous schema. In this paper, we propose a transductive hybrid approach composed of an unsupervised node representation learning model followed by a node classification/Edge prediction model. The proposed model is capable of processing heterogeneous graphs to produce unified node embeddings which are then utilized for node classification or link prediction as the downstream task. The proposed model is developed to classify stock market technical analysis reports, which to our knowledge is the first work in this domain. Experiments, which are carried away using a constructed dataset, demonstrate the ability of the model in embedding extraction and the downstream tasks. 
\end{abstract}

\maketitle
\pagestyle{plain}

\section{Introduction}


The way we understand and process natural language mainly depends on the way it is represented. While representing language as a bag of tokens and sequence of tokens are largely used in the NLP community, graph representation is another approach for structuring texts, as well as utilizing relational information between different text elements. Representing text as a graph has a long history as several specialized text graphs, including dependency graphs \cite{zhang2019aspect}, constituency graphs \cite{marcheggiani2019graph}, lexical networks \cite{radev2008networks}, knowledge graphs \cite{ye2019vectorized}, etc. has been created in this field. Moreover, it is also possible to represent entities with different hierarchies such as document, passage, sentence, and word in one graph. Because of this property, many NLP tasks require heterogeneous graphs to model different entities and relations of the problem. However, GNN models, as one of the most powerful frameworks of analyzing graphs, such as GCN\cite{kipf}, GAT\cite{velivckovic2017graph}, and GraphSAGE\cite{hamilton2017inductive} are mostly designed for homogeneous conditions, and therefore cannot fit in these problems directly. 
\\
We identify stock market technical analysis reports as a text dataset that provides great analytical opportunities in financial markets. Hundreds of these analyses in the form of text documents are published in TradingView\footnote{Tradingview.com} on a daily basis. Market analysts and investors share their analysis, insights, and predictions within the platform. They can also assign a tag to their posts from a predefined set (Long, Short, Education) to indicate the overall status of that analysis. Table \ref{tab:sampledata} shows a sample of a post in our dataset.
\\
As a novel work in this domain, we strive to automatically extract insights and information from these reports with the help of graph representation learning. In particular,  this paper focuses on a specific version of that vision that is the problem of classifying technical analysis reports in the cryptocurrencies market. Classifying all documents into the Long, Short, and Education classes helps us to perceive what position analysts and investors generally take on a particular symbol in a given time window. In order to classify properly, we use authors' tags as the supervision of the classification task and train a classifier based on those labels. 

\begin{table}
    \centering
    \caption{Sample of the dataset}
    \label{tab:sampledata}
    \begin{tabularx}{\columnwidth}{c X}
    \toprule
    \textbf{Feature} & \textbf{Sample}\\
    \midrule
    \textbf{author} & EXCAVO \\
    \textbf{\#comments} & 7 \\
    \textbf{content} & ZECUSDT   reached 50\% Fibonacci and forming a  bullish flag...\\ 
    \textbf{ID} & L68rv5WO \\ 
    \textbf{\#likes} & 65 \\ 
    \textbf{position} & Long \\ 
    \textbf{signature} & Channels: https://t.me/excavochannel ... \\ 
    \textbf{symbol} & BINANCE:ZECUSDT \\
    \textbf{time} & Jan 21, 2021 @ 17:21:24.000 \\
    \textbf{timeframe} & 240 \\
    \textbf{title} & ZEC/USDT/BTC \\
    \bottomrule
    \end{tabularx}

\end{table}

To exploit richer relations among documents, we follow the graph representation approach for the proposed document classification problem and construct a graph based on the given input text. There are multiple methods for static graph construction during preprocessing documents \cite{wu2021graph}. We take a hybrid approach, in which we capture both co-occurrence and similarity relations between words and documents. We also cover the loss of the text sequential information to some extent, by creating n-grams during the preprocessing steps and using them in graph modelling.
\\
Our base graph structure is solely constructed on the information contained in the technical reports: documents and words, but formulating this classification problem in a graph-based structure allows us to represent our desired entities that are related to our specific problem as part of the graph structure. We can take price patterns as an example, which are formations that appear on stock price charts and are one of the key components of technical analysis in financial markets. 
If we consider each price pattern as a separate node type, it is possible to connect each one to its implication defined by the tag node type, as well as the document containing that pattern. Here, we have used an external knowledge base to enrich our document dataset. As a result, we can incorporate different domain knowledge into the current structure, and capture richer relationships among different elements.
\\
In this work, we introduce a novel classification task of stock market technical analysis documents and develop a hybrid transductive graph-based solution.  In the proposed approach, first, we store TradingView’s posts published in the Ideas section and preprocess all documents. Then, we define word and sentence representations by language models to construct our graph. We also add price patterns, corpus topics, and post labels to the graph as separate node types. After constructing the graph, node embeddings are trained in an unsupervised manner, and we update the graph nodes with new representations. This step can be considered optional. Finally, we train both node classification on documents and link prediction on the edges between documents and tags in order to estimate document labels. We conduct various experiments to examine our proposed method and compare it with state-of-the-art baselines in different settings. Experimental results show that our best graph-based approach achieves 89\% F1-score, while the top baseline method (Fine-tuned BERT) has the F1-score 76\%.
\\
In short, we summarize the main contributions as follows: 
\begin{itemize}

    \item Proposing a novel heterogeneous graph construction method for representing technical analysis documents.
    
    \item Proposing a graph neural network model for technical analysis reports classification. 
    
    
    \item Providing experimental results showing that our method outperforms several state-of-the-art document classification baselines. 
    
    \item Introducing the ``Stock Market Technical Analysis Reports Classification” task, which to our knowledge is a novel work in related domains.
    
    \item Presenting a dataset of technical analysis reports in the cryptocurrencies market that enables further research in this field.
\end{itemize}


The remaining parts of this paper is structured as follows. Section \ref{sec:2} further discusses the stock market technical analysis reports and price patterns in financial markets. Following that, in Section \ref{sec:3}  our graph construction approach is described and our hybrid proposed model is explained in detail. Section \ref{sec:4} introduces the baseline methods, our constructed dataset, and experimental results. Finally, a categorization of graph-based approaches for text classification is presented as related works in Section \ref{sec:5} and the paper is concludeed in Section \ref{sec:6}.

\section{Stock Market Technical Analysis Report}\label{sec:2}
Predicting market behaviour from financial text data has been under extensive research in recent years. Analyzing financial texts everyday and extracting insights from them is a heavy task, therefore, researchers have been trying to leverage machine learning methods to derive insights. Research in this area has been mostly done on sentiment analysis of social media data to find out whether people are talking about the market in a positive, negative or neutral way \cite{finbert, CHAN201753, sentimentfin, sfin2, 7752381, DANIEL2017111, 9437616, 8334488}. These approaches miss an important point. Most of the times a rise or fall in market prices leads to social media posts and people's reactions. It is rare to get insights about market's future trend from people's emotions in their posts. 
To address this matter, technical analysis reports are used in this study. Technical analysts use price movements in price charts to decide the future trends of the price \cite{10.5555/1405755}. They use different indicators and patterns for their analysis. Technical analysis reports are the outcome of technical analysts' investigation of the price chart. They talk about the behaviour of the price chart and how they think the behaviour affects the future trend of the chart. These reports carry much more information about the movements of the market. Therefore, it is more likely that technical reports are correlated with price movements because writers are usually traders with expertise in market analysis.


\subsection{Research Target}
Market analysts usually talk about their analysis of the market behaviour and how they think market will behave afterwards in their writings. In this study, we aim to process their analysis and find out what they expect to see in the chart. However, technical analysis can be prone to error as not every analysis is correct and not everyone is expert enough to predict future behaviour correctly. By looking at what the majority of the people are seeing in the price chart and implications in the market, it is possible to derive insights from their implications and finally predict the possible next move of the market from the technical analysts' point of view.

\subsection{Price Pattern}
Price patterns are configurations in price charts that are often used by technical analysts for anticipating the direction of the market prices. Price patterns have different shapes and are often identified by trendlines and curves \cite{seo2017scientific}. Patterns can be categorized into two groups of reversal and continuation patterns. Reversal patterns state a change in the direction of the market and continuation patterns state that the trend remains the same. In this work, most popular price patterns among technical analysts were used as graph nodes. In stock trading, having a long position means that a rise is expected in the price of the market as opposed to a short position which means a fall is expected in the price. Price patterns usually signal a rise or fall afterwards so their implications in the market is utilized in this research in graph modeling. Table \ref{tab:patterns} shows some examples of patterns, their description and implications in the market.
\begin{table}
    \centering
    \caption{Price patterns}
    \label{tab:patterns}
    \begin{tabularx}{\columnwidth}{p{0.15\linewidth} X c}
    \toprule
     \textbf{Name} & \textbf{Description} & \textbf{Implication}\\
     \midrule
     Head and shoulders & \small{A baseline with three peaks where the middle is the highest.} & Short\\
     Double top & \small{Two highs, looking like M} & Short \\ 
     Cup and handle & \small{Two convex arcs which three top points touch a resistance level} & Long\\
     Bearish flag & \small{Characterized by parallel trendlines over the consolidation area-lines have upward sloping} & Short\\
     \bottomrule
    \end{tabularx}
\end{table}

\section{Proposed Approach}\label{sec:3}
In this section, our approach for financial text analysis is explored. First, the graph structure is explained and then graph processing approach is introduced in details. Figure \ref{fig:overview} illustrates the overall process.


\tikzstyle{database} = [cylinder, minimum width=1.5cm, minimum height=1.5cm,text centered, draw=black, rotate=90 ,fill=black!5]
\tikzstyle{process} = [rectangle, minimum width=3cm, minimum height=1cm, text centered, text width=3cm, draw=black, fill=black!5]
\tikzstyle{dashedprocess} = [draw=gray,dashed,fill=black!5,thick,inner sep=5pt,minimum width=3cm, minimum height=1cm, text width=3cm, text centered]
\tikzstyle{arrow} = [thick,->,>=stealth]
\tikzstyle{line} = [thick,-]

\begin{figure*}[h]

    \centering
    
\scalebox{0.65}{
\begin{tikzpicture}[node distance=4 cm]
\node (db) [database] {DataBase};
\node (pp) [process, right of=db] {Text\\ Pre-processing};
\node (tm) [process, above of= pp, yshift=-1.5cm] {Topic Modeling};
\node (te) [process, right of=pp] {Text Embedding};
\node (gm) [process, right of=te] {Graph Modeling};
\node (nu) [coordinate, right of = gm, xshift=-1cm]{};
\node (nc) [process, above of=nu, xshift=2.5cm, yshift=-3.2cm] {Node Classification};
\node (lp) [process, below of=nu, xshift=2.5cm, yshift=3.2cm] {Link Prediction};
\node (dgi) [dashedprocess, below of = gm, yshift=0.5cm] {Unsupervised Node Representation Learning};
\node (nu2) [coordinate, right of = dgi, xshift=-1cm]{};

\node (nc2) [dashedprocess, above of=nu2, xshift=2.5cm, yshift=-3.2cm] {Node Classification};
\node (lp2) [dashedprocess, below of=nu2, xshift=2.5cm, yshift=3.2cm] {Link Prediction};

\draw [arrow] (db) -- (pp);
\draw [arrow] (pp) -- (tm);
\draw [arrow] (tm) -| (gm);
\draw [arrow] (pp) -- (te);
\draw [arrow] (te) -- (gm);
\draw [line] (gm) -- (nu);
\draw [arrow] (nu) |- (nc);
\draw [arrow] (nu) |- (lp);
\draw [arrow] (gm) -- (dgi);
\draw [line] (dgi) -- (nu2);
\draw [arrow] (nu2) |- (nc2);
\draw [arrow] (nu2) |- (lp2);
\end{tikzpicture}
}
\caption{Overview of the proposed model}
\label{fig:overview}
\end{figure*}
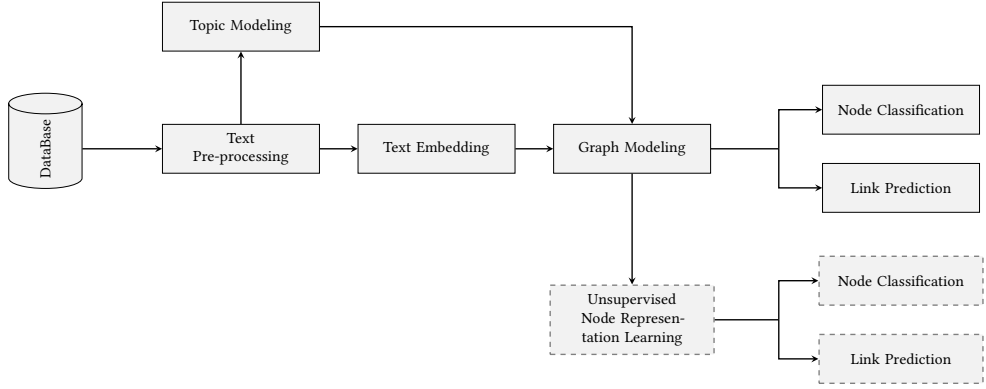

\subsection{Graph Structure}
In this section, details of our text graph construction approach are explained. Every graph \(\mathcal{G}(\mathcal{V}, \mathcal{E})\) is characterised by a set of nodes and edges. Our graph's set of nodes includes five types of nodes:
\begin{itemize}
    \item Documents
    \item Words
    \item Topics
    \item Price patterns
    \item Document position (Link Prediction problem)
\end{itemize}
Document and word nodes are basics for representing a corpus
as a graph and other types of nodes are added to enrich the
graph representation. Document position nodes are labels of the documents that are only used in the link prediction problem where the existence of a link between each document and position nodes are predicted. 

\begin{table}
\caption{Description of edges of the graph}
\label{tab:edges}
\begin{tabularx}{\columnwidth}{c c X}
\toprule
 \textbf{Source} & \textbf{Destination} & \textbf{Relation} \\ 
\midrule
  Word & Word & PMI \\

  Document & Word & TF-IDF \\

  Word & Topic & $P(w|t)$ \\

  Document & Topic  & $P(t|d)$\\ 

  Document & Price Pattern  & binary\\

  Price Pattern & Topic & $P(t|pp)$ \\

 Document & Position & User-generated label (Link Prediction problem) \\
 
 Price Pattern & Position & Price pattern implication (Link Prediction problem) \\
\bottomrule
\end{tabularx}

\end{table}

\tikzstyle{arrow} = [very thick,->,>=stealth]
\tikzstyle{line} = [very thick, -]
\tikzstyle{dline} = [very thin, -]

\tikzstyle{word} =  [circle, draw, very thick, fill=green!50, text width=1cm, text centered]
\tikzstyle{pattern} =  [circle, draw, very thick, fill=blue!50, text width=1cm, text centered]
\tikzstyle{topic} =  [circle, draw, very thick, fill=red!50, text width=1cm, text centered]
\tikzstyle{document} = [rectangle, very thick, rounded corners, draw=black, fill=orange!50, text width=4cm]

\tikzstyle{dummynode} =  [circle, draw,  fill=black!5, text width=1cm, text centered]
\tikzstyle{dummydocument} = [rectangle, rounded corners, draw=black, fill=black!5, text width=4cm]

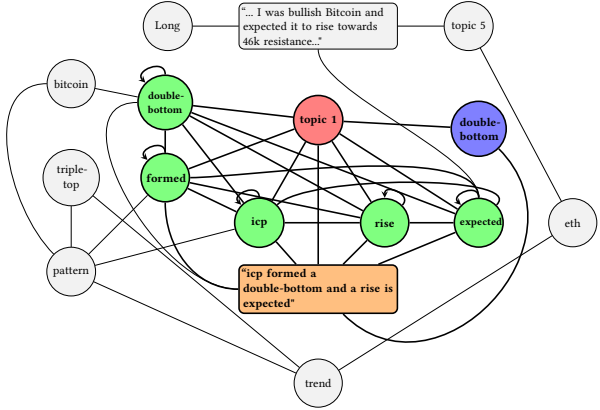
\begin{figure}[t]
\centering
\scalebox{0.5}{
\begin{tikzpicture}[node distance=2.5cm]
\node (doc) [document] {\textbf{``icp formed a\\ double-bottom and a rise is expected"}};

\node (word1) [word, above left of = doc, xshift=0.2cm] {\textbf{icp}};
\node (word2) [word, left of = word1, yshift=1.2cm] {\textbf{formed}};
\node (word3) [word, above of = word2, yshift=-0.5cm] {\textbf{\small{double-bottom}}};

\node (topic) [topic, above of = doc, yshift=2cm] {
\textbf{topic 1}};

\node (dd) [dummydocument, above of = topic] {``... I was bullish Bitcoin and expected it to rise towards 46k resistance..."};
\node (dtopic) [dummynode, right of = dd, xshift=1.5cm]{topic 5};
\node (dposition) [dummynode, left of = dd, xshift=-1.5cm]{Long};
\node (dword2) [dummynode, below of = doc]{trend};
\node (dpp) [dummynode, left of = word2] {triple-top};
\node (dword4) [dummynode, below of = dpp] {pattern};

\node (word4) [word, above right of = doc] {\textbf{rise}};
\node (word5) [word, right of = word4] {\textbf{\small{expected}}};
\node (pattern) [pattern, above of = word5] {\textbf{double-bottom}};

\node (dword3) [dummynode, right of = word5]{eth};
\node (dword1) [dummynode, above of = dpp]{bitcoin};

\draw [line] (doc) -- (word1);
\draw[line] (doc) to [out=180, in=270, looseness=1] (word2);
\draw (doc) to [out=180, in=180, looseness=1] (word3);
\draw [line] (doc) -- (word4);
\draw [line] (doc) -- (word5);

\draw [line] (doc) to [out=315, in=315, looseness=1.5] (pattern);

\draw [line] (topic) -- (word1);
\draw [line] (topic) -- (word2);
\draw [line] (topic) -- (word3);

\draw [line] (topic) -- (word4);
\draw [line] (topic) -- (word5);
\draw [line] (topic) -- (doc);
\draw [line] (topic) -- (pattern);

\draw[arrow] (word1) to [out=90, in=135, looseness=2] (word1);
\draw[arrow] (word2) to [out=90, in=135, looseness=2] (word2);
\draw[arrow] (word3) to [out=90, in=135, looseness=2] (word3);
\draw[arrow] (word4) to [out=45, in=90, looseness=2] (word4);
\draw[arrow] (word5) to [out=90, in=135, looseness=2] (word5);

\draw[line] (word1) -- (word2);
\draw[line] (word1) -- (word3);
\draw[line] (word1) -- (word4);
\draw[line] (word1) to [out=45, in=45, looseness=0.5] (word5);
\draw[line] (word2) -- (word3);
\draw[line] (word2) to [out=0, in=90, looseness=0.5] (word5);
\draw[line] (word2) -- (word4);
\draw[line] (word3) -- (word4);
\draw[line] (word3) -- (word5);
\draw[line] (word4) -- (word5);

\draw[dline] (dd) -- (dtopic);
\draw[dline] (dd) -- (dposition);
\draw[dline] (dd) to [out=270, in=90, looseness=0.5] (word5);
\draw[dline] (dword1) -- (word3);

\draw[dline] (dword2) -- (dword3);
\draw[dline] (dword3) -- (dtopic);
\draw[dline] (dword4) -- (dpp);
\draw[dline] (dword2) -- (dpp);

\draw[dline] (word1) -- (dword4);
\draw[dline] (word2) -- (dword4);
\draw[dline] (dword4) to [out=135, in=180, looseness=1] (dword1);
\draw[dline] (dword2) -- (dword4);

\end{tikzpicture}
}
\caption{An illustration of sample graph representation }
\label{fig:samplegraph}
\end{figure}

We connect words and documents based on word occurrence in documents so that the weight of the edges are defined by term frequency-inverse document frequency (TF-IDF) \cite{yao2019graph}. TF-IDF evaluates how relevant a word is to a document in the entire document set. Term frequency is the number of times the word appears in the document and inverse document frequency is the inverse of the number of documents that contain that word. 
\\
We use a fixed-size sliding window on all documents to calculate point-wise mutual information (PMI) between every two words within the window. Then, we connect words with positive PMI values together and assign the derived PMI as the weight of that connection. The PMI value of a word pair \(w_1, w_2 \) is computed as

\begin{equation}
\operatorname{PMI}(w_1, w_2)=\log \frac{p(w_1, w_2)}{p(w_1) p(w_2)}
\end{equation}

\begin{equation}
p(w_1, w_2)=\frac{\# W(w_1, w_2)}{\# W}
\end{equation}

\begin{equation}
p(w_1)=\frac{\# W(w_1)}{\# W}
\end{equation}
Where \(\# W(w_1)\) is the number of sliding windows in a corpus that contain word \(w_1\), \(\# W(w_1, w_2)\) is the number of sliding windows that contain both \(w_1\) and \(w_2\), and \(\#W\) is the total number of sliding windows. In this approach, we have used both co-occurrence and similarity relations between words and documents. Each of those methods only considers one specific type of relation and has limitations in representing others. 
\\
To utilize high-level semantic information of documents, we connect topic node \(t\), resulted from the topic modeling, to document, word, and price pattern nodes with the relations shown in the table \ref{tab:edges}. As aforementioned, we introduce the price pattern node type that is connected to the documents containing that pattern. We also connect patterns that have known implications in the technical analysis literature to the position nodes.
Figure \ref{fig:samplegraph} illustrates a sample of nodes and their relations in our graph. Table \ref{tab:edges} summarizes edges between different node types.


\subsection{Heterogeneous Graph Modeling}
As we are dealing with a heterogeneous structure, we design a model that is capable of processing these kinds of graphs. In this section, we describe the components of our proposed model. 

\subsubsection{\textbf{HinSAGE}}
GraphSAGE \cite{hamilton2017inductive} is a framework for inductive representation learning on large graphs. GraphSAGE is used to generate low-dimensional vector representations for nodes, and is capable of learning an embedding function that generalizes to unseen nodes, without requiring a re-training procedure. 
The key idea of this approach is training a set of aggregator functions that learn to aggregate feature information from a node’s neighborhood. 
\newline
We extend the GraphSAGE model to be able to aggregate feature vectors from different node types in heterogeneous graphs. Our work is inspired by the HinSAGE model introduced by the StellarGraph library \cite{StellarGraph}. Algorithm \ref{alg:graphsage} describes HinSAGE in the case where entire graph \(\mathcal{G}(\mathcal{V}, \mathcal{E})\) and features for all nodes are provided as input. First, all node embeddings are initialized to the node features. Then, at each iteration (search depth), each node aggregates the representations of the nodes in its one-hop neighborhood via edges of type \(r\), and concatenates its current representation \(\mathbf{h}_{v}^{k-1}\) with the aggregated neighborhood vector, \(\mathbf{h}^{k}_{N_{r}(v)}
\). As we can observe in the algorithm, there are separate neighborhood weight matrices \(W_{\text{neigh}}\) for each relation between two node types. There are also separate self-feature matrices \(W_{\text{self}}\) for every node type. HinSAGE finally passes the concatenated vector through a neural network layer with non-linear activation function \(\sigma\) to update the node embedding.

\begin{algorithm}

\caption{HinSAGE Algorithm}\label{alg:graphsage}

\hspace*{\algorithmicindent} \textbf{Input:}  
Graph \(\mathcal{G}(\mathcal{V}, \mathcal{E})\); input features \(\left\{\mathbf{x}_{v}, \forall v \in \mathcal{V}\right\}\); edge types \(r \in \{1, \ldots, R_e\}\); depth \(K\); weight matrices \(W^{k}, \forall k \in\{1, \ldots, K\}\); non-linearity \(\sigma\); neighborhood function \(N: v \rightarrow 2^{\mathcal{V}}\)\\

\hspace*{\algorithmicindent} \textbf{Output:} Vector representations \(\mathbf{z}_v\) for all \(v \in \mathcal{V}\)

\begin{algorithmic}[1]
\State $\mathbf{h}_{v}^{0} \gets \mathbf{x}_{v}, \forall v \in \mathcal{V} ;$
\For {$k=1 \ldots K $}
\For {$ v \in \mathcal{V} $}
\For {$r=1 \ldots R_e$}
\State ${\mathbf{h}^{k}}_{N_{r}(v)} \gets \frac{1}{|N_{r}(v)|}\sum_{u \in N_r(v)}D_{p}\lbrack{\mathbf{h}_{u}}^{k - 1}\rbrack
 $
\EndFor

\State ${\mathbf{h}_{v}}^{k} \gets \sigma \lbrack \operatorname{CONCAT}\lbrack{W^{k}}_{t_{v},\text{self}}D_{p}\lbrack{\mathbf{h}_{v}}^{k - 1}\rbrack, {W^{k}}_{1,\text{neigh}} {\mathbf{h}^{k}}_{N_{1}(v)},$\newline \hspace*{\algorithmicindent}
\hspace*{\algorithmicindent}\hspace*{\algorithmicindent}\hspace*{\algorithmicindent}\hspace*{\algorithmicindent}\hspace*{\algorithmicindent}$\ldots, {W^{k}}_{R_{e},\text{neigh}}{\mathbf{h}^{k}}_{N_{R_{e}}(v)}\rbrack + b^{k} \rbrack$

\EndFor
\State $\mathbf{h}_{v}^{k} \leftarrow \mathbf{h}_{v}^{k} /\left\|\mathbf{h}_{v}^{k}\right\|_{2}, \forall v \in \mathcal{V}$
\EndFor
\State $\mathbf{z}_{v} \leftarrow \mathbf{h}_{v}^{K}, \forall v \in \mathcal{V}$
 
\end{algorithmic}
\end{algorithm}

In order to learn the weights of the aggregator functions, GraphSAGE applies a differentiable, graph-based loss function: 

\begin{equation}
J_{\mathcal{G}}\left(\mathbf{z}_{u}\right)=-\log \left(\sigma\left(\mathbf{z}_{u}^{\top} \mathbf{z}_{v}\right)\right)-Q \cdot \mathbb{E}_{v_{n} \sim P_{n}(v)} \log \left(\sigma\left(-\mathbf{z}_{u}^{\top} \mathbf{z}_{v_{n}}\right)\right),
\end{equation}

where node \(v\) is a neighbor of node \(u\), node \(v_n\) is a distant node to node \(v\) and is sampled from a negative sampling distribution \(P_{n}(v)\), and \(Q\) is the number of negative samples. This loss function enforces close nodes to have similar representations and distant nodes to have dissimilar representations.

\subsubsection{\textbf{DGI}}
Deep Graph Infomax (DGI) \cite{velickovic2019deep} is an unsupervised node representation learning approach for graph-structured data. DGI's core idea is learning to distinguish between the original graph and the corrupted one that derives from a corruption procedure. Given the true input graph \(G\), they change it into the mutated version \(H = C(G)\) by randomly shuffling the node features among nodes. \(H\) has the same edges as \(G\), but the features associated with each node differ.
\\
The model consists of an encoder that takes an input graph and computes an embedding vector for each node. It is typically based on well-known graph machine learning models such as GCN or GraphSAGE. The weights of the encoder are trained according to the distinguishing step. Thus, DGI learns to discriminate between nodes that have sensible connections and nodes that have unexpected connections. After training, the encoder can be used independently to compute node embedding vectors directly. We use DGI to obtain node embeddings for the downstream tasks as shown in the figure \ref{fig:overview}. Because of heterogeneity, we separately run the same algorithm with the same parameters for each node type.

\section{Experiments and Results}\label{sec:4}
\subsection{Dataset}
 Tradingview is a platform and social network for traders and investors where they can use different charts and technical features to make trading decisions. Traders can also share their opinions and analysis of the market in a section called Ideas. Users can choose between three options (i.e., Long, Short, and Education) to describe general purpose of their post. These user-generated labels were utilized for text classification tasks and a dataset was put together from people's posts about the cryptocurrency market and their decision inferred from their analysis.

In this work, posts from 163 most popular cryptocurrency symbols were collected and stored in a database. In addition to assigning a label, market analysts select a timeframe for their posts to define the time range where their analysis or decisions are valid. Each post has a signature part in which writers usually share their websites or Telegram channels. Table \ref{tab:sampledata} shows all the crawled information from Tradingview Ideas and a sample data point.\
Posts containing no text data were discarded. Overall 24420 posts are available in the dataset. Table \ref{tab:statistics} shows some statistics of our dataset.
\begin{table}
    \centering
    \caption{Statistics of the dataset}
    \label{tab:statistics}
    \begin{tabularx}{\columnwidth}{X c}
    \toprule
    \textbf{Field} & \textbf{Amount}\\
    \midrule
    \textbf{Total posts} & 24420\\
    \textbf{Total labelled posts} & 16590\\
    \textbf{Mean length of posts} & 70.05 words\\
    \textbf{Median length of posts} & 34 words\\
    \textbf{Total posts containing at least one price pattern} & 793\\
    \textbf{Posts labelled as Long} & 13379(80.6\% of labels)\\
    \textbf{Posts labelled as Short} & 2984(18\% of labels)\\
    \textbf{Posts labelled as Education} & 227(1.4\% of labels)\\
    \bottomrule
    \end{tabularx}

\end{table}

\subsection{Pre-process}

Text data needs to be normalized and preprocessed for further use. In order to clean the documents in the corpus, these steps were taken:
\begin{itemize}
    \item Normalizing documents
    \item Removing stop-words and punctuation marks
    \item Removing email addresses, newline characters, URLs and emojis
    \item Unifying price pattern variations, e.g., changing \textit{head n shoulders}, \textit{head \& shoulders}, \textit{h \& s} to \textit{head-and-shoulders}.
\end{itemize}
After cleaning the documents, they were broken down into their words to form a dictionary of words from the corpus.\newline

\subsection{Topic Modeling}

\begin{figure}
\centering
\scalebox{0.7}{
\begin{tikzpicture}
\begin{axis}[
    xlabel={Number of topics},
    ylabel={Coherence score},
    xmin=3.5, xmax=10.5,
    ymin=0.39, ymax=0.46,
    xtick={4,5,6,7,8,9,10},
    ytick={0.40,0.41,0.42, 0.43, 0.44,0.45,0.46},
    legend pos=north east,
    ymajorgrids=true,
    xmajorgrids=true,
    grid style=dashed,
    enlargelimits=false
]

\addplot[
    color=blue,
    mark=square,
    ]
    coordinates {
    (4,0.4504)(5,0.4367)(6,0.450488)(7,0.425672)(8,0.418337)(9,0.397188)(10,0.4111)
    };
    \legend{score}
    
\end{axis}
\end{tikzpicture}
}
\caption{Selecting the suitable number of topics}
\label{fig:topicselection}
\end{figure}
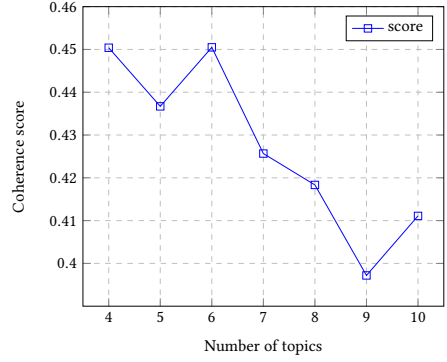

As stated in previous sections, topic nodes are part of our graph structure. To extract topics and their distribution, Latent Dirichlet Allocation (LDA) method \cite{lda} is utilized. To find an optimal number of topics for the corpus, coherence score was calculated for different topic numbers. Figure \ref{fig:topicselection} shows the result. This measure evaluates a single topic by measuring the degree of semantic similarity between high scoring words in the topic. This kind of measurements help
distinguish between topics that are semantically interpretable topics and topics that are artifacts of statistical inference. There are various ways to calculate coherence score. Here we have used the Context Vector coherence score introduced by \cite{Aletras2013EvaluatingTC} that uses words co-occurrence counts. Based on the results shown in Table \ref{fig:topicselection} the optimal number of topics was set to 6. 

\begin{figure}
     \centering
     \begin{subfigure}[b]{0.22\textwidth}
         \centering
         \includegraphics[width=\textwidth]{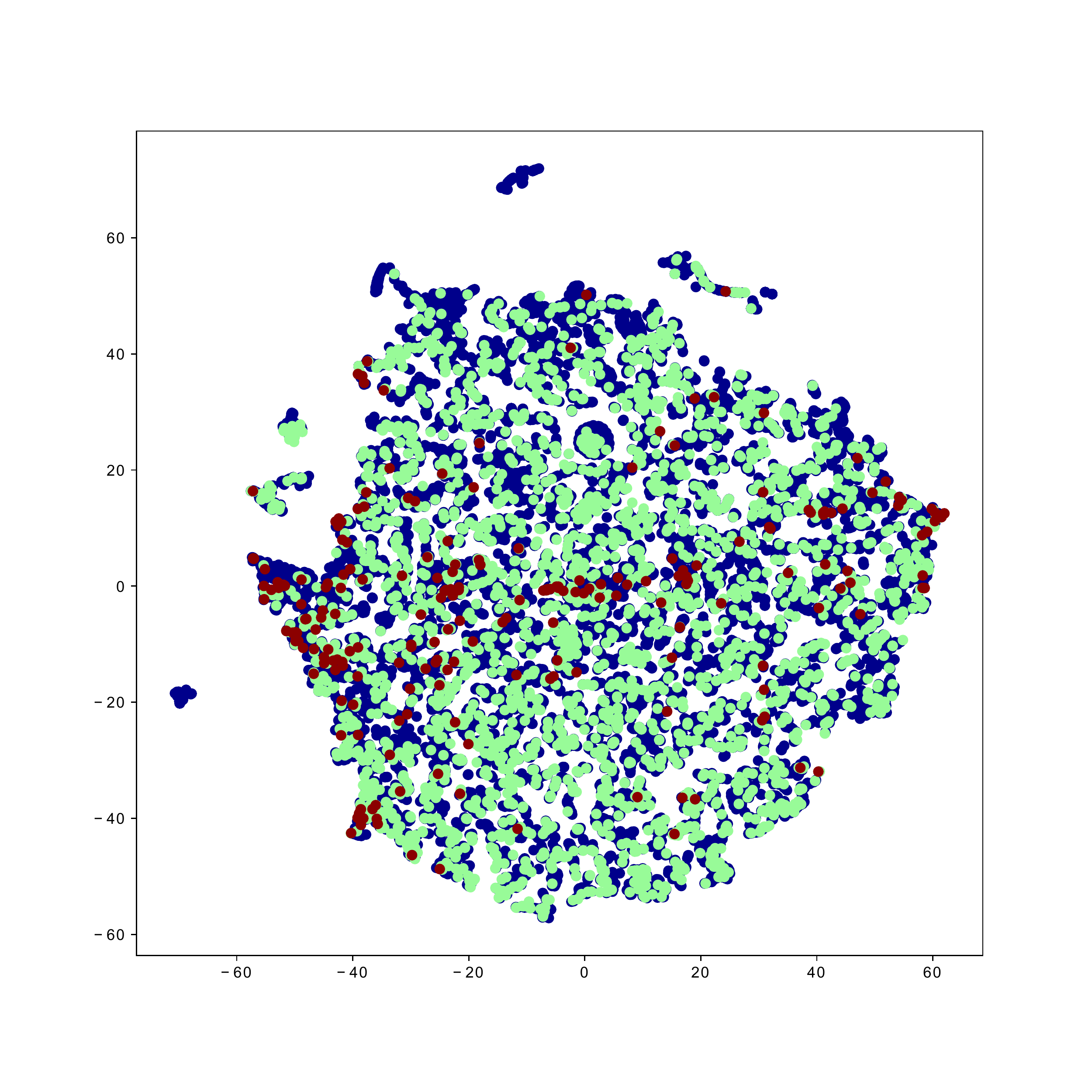}
         \caption{}
         \label{fig:topic}
     \end{subfigure}
     \hfill
     \begin{subfigure}[b]{0.22\textwidth}
         \centering
         \includegraphics[width=\textwidth]{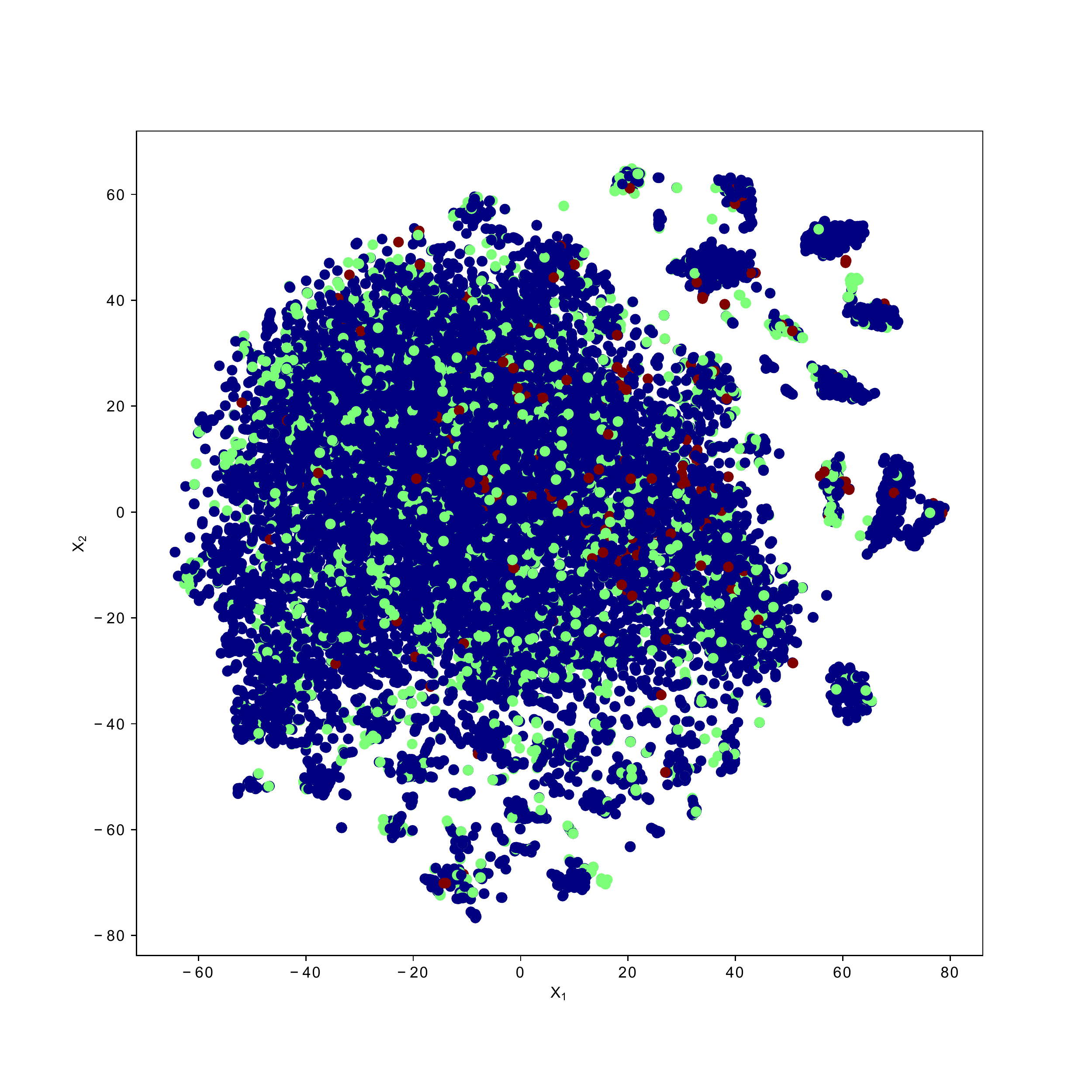}
         \caption{}
     \end{subfigure}
     \\
     \medskip
     \begin{subfigure}[b]{0.22\textwidth}
         \centering
         \includegraphics[width=\textwidth]{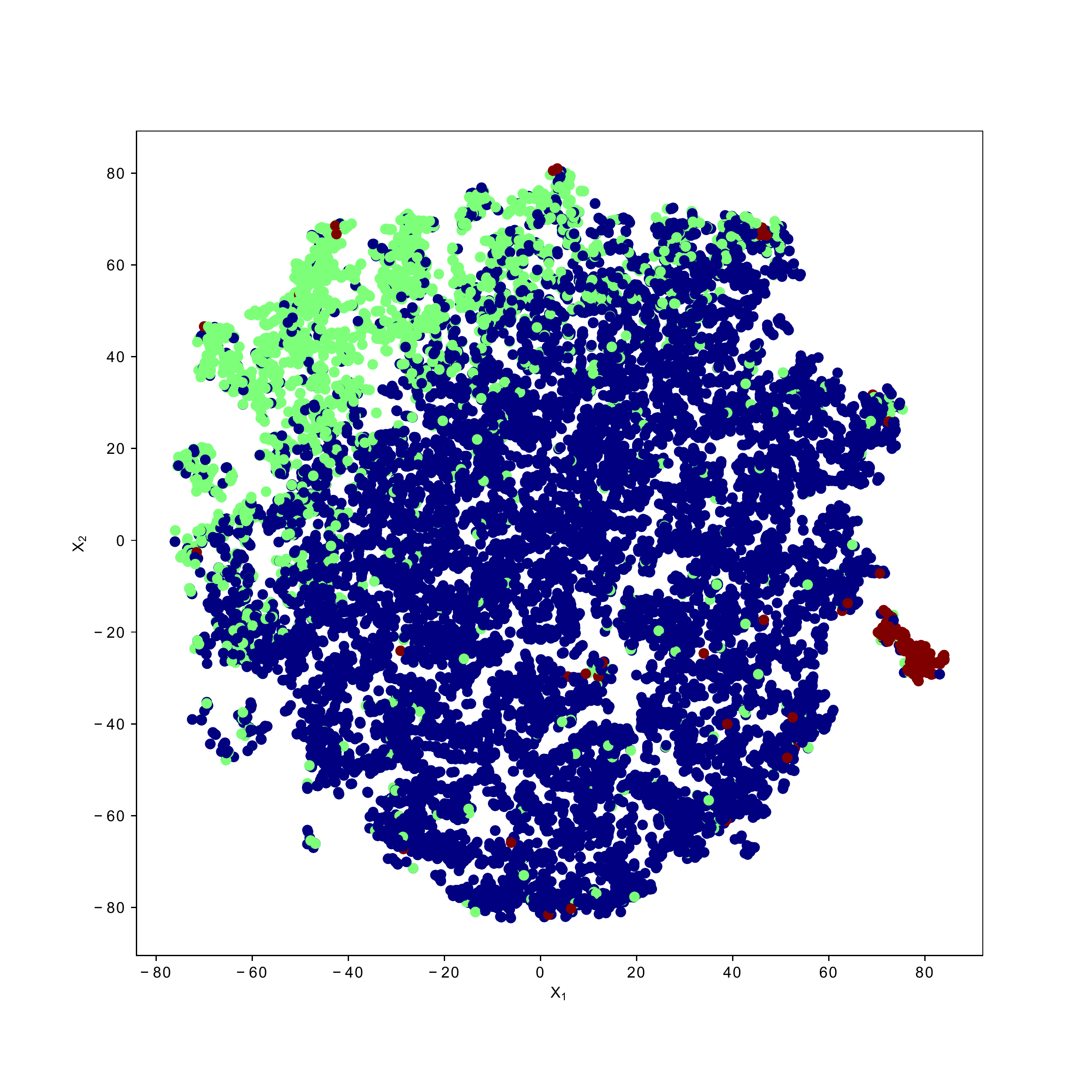}
         \caption{}
     \end{subfigure}
        \hfill
     \begin{subfigure}[b]{0.22\textwidth}
         \centering
         \includegraphics[width=\textwidth]{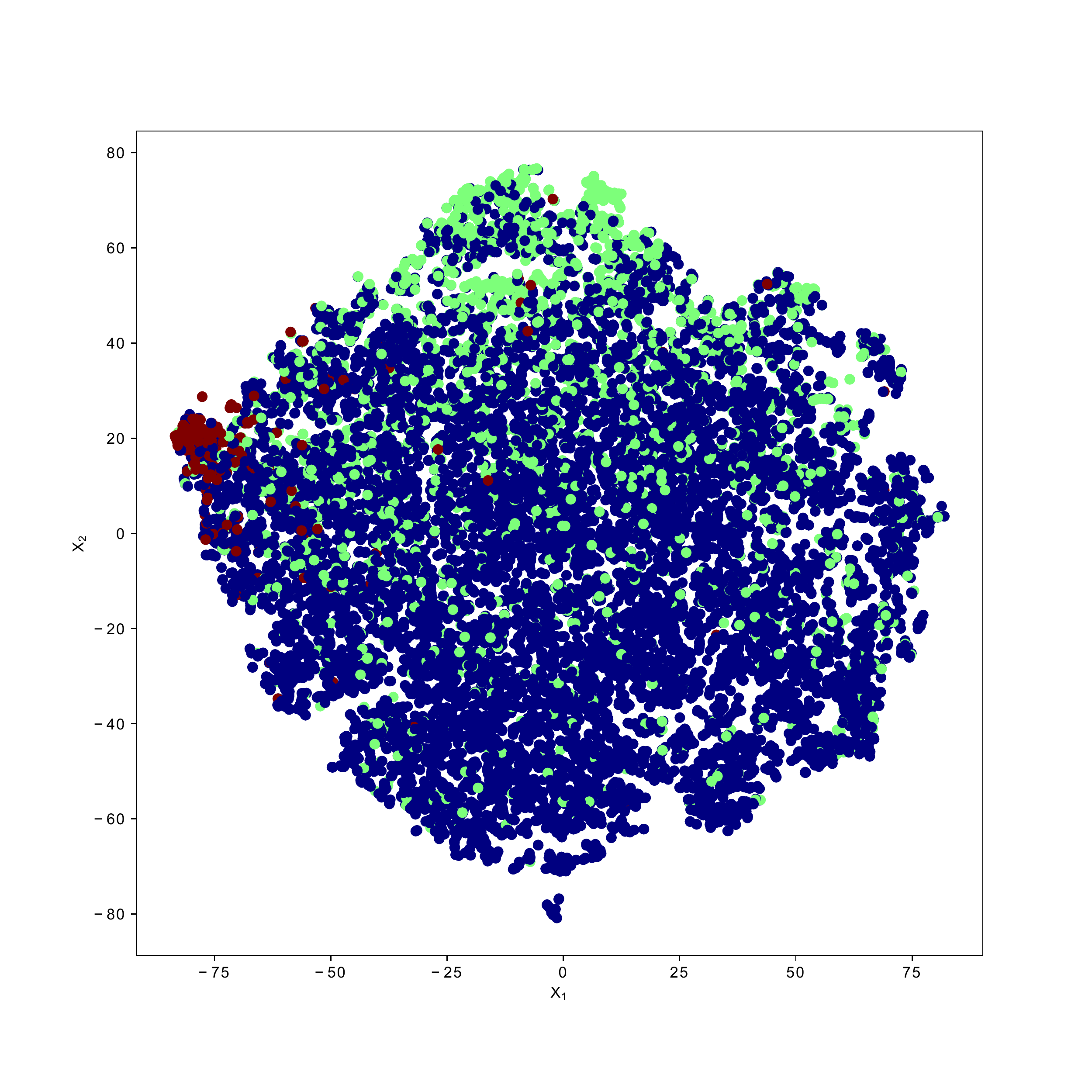}
         \caption{}
     \end{subfigure}
    \caption{Embedding of documents in 4 different settings. (a): embeddings derived from topic distributions, (b): embeddings that are learned by DGI, (c): embeddings learned through HinSAGE node classification, and (d): embeddings learned through HinSAGE node classification on DGI representations. The blue, green, and red dots correspond to the documents with Long, Short, and Education labels respectively.}
    \label{fig:hinsage-nc}
\end{figure}

The final results of the topic modeling is shown in Table \ref{tab:topics} in the form of each topic's most frequent words. Each topic's distribution was calculated for each document. Then the resulting vectors were mapped into two-dimensional space using t-SNE method \cite{JMLR:v9:vandermaaten08a}. The visualization of the data-points is shown in Figure \ref{fig:hinsage-nc}. The distributions of words in topics and topics in documents is later used in our modeling process. The effectiveness of topic nodes in classification task is studied and the results show that these nodes improve the accuracy of the prediction.

\begin{table}
\centering
    \caption{Topics and their frequent words}
    \label{tab:topics}
    \begin{tabularx}{\columnwidth}{ c c c c c c }
        \toprule
         \textbf{1} & \textbf{2} & \textbf{3} & \textbf{4} & \textbf{5} & \textbf{6} \\
        \midrule
        \small{time} & \small{buy} & \small{binance} & \small{drop} & \small{bch} & \small{price}\\
    
        \small{bitcoin} & \small{term} & \small{analysis} & \small{trendline} & \small{crypto} & \small{support}\\
    
        \small{market} & \small{tp} & \small{coin} & \small{beginning} & \small{structure} &\small{btc}\\
    
        \small{wave} & \small{trade} & \small{position} & \small{monthly} & \small{charts} &\small{bullish}\\
    
        \small{zec} & \small{short} & \small{dont} & \small{find} & \small{startegy} & \small{long}\\
        
        \bottomrule
    \end{tabularx}
\end{table}

For initial embedding of graph nodes, two models were used. FastText library \cite{fasttext} was used to encode words in our dictionary in a 100 dimensional embedding space. To encode the documents, sentence-transformers \cite{sentence-bert} pre-trained MiniLM  model\cite{minilm} with 6 layers was used. The MiniLM model maps documents to a 384 dimensional embedding space. Summary of all node embeddings and their dimensions are shown in Table \ref{tab:embeddings}. 
\begin{table}
\caption{Graph nodes' initial embedding}
\label{tab:embeddings}
\begin{tabularx}{\columnwidth}{c X c}
\toprule
 \textbf{Nodes} & \textbf{Embedding} & \textbf{Dim} \\ 
 \midrule
 Word node &  \textit{fasttext} word embedding & 100 \\

 Document node & Pretrained LM document embedding & 384\\ 

 Topic node & One-hot & 6 \\

 Price Pattern node & One-hot & 29 \\ 

 Position node & One-hot (used in the Link Prediction problem.) & 3  \\ 
\bottomrule
\end{tabularx}
\end{table}

\subsection{Baselines and Results}
The following models are used as baselines for comparison purposes:
\begin{itemize}
    \item A classification head was added to the pre-trained uncased BERT base model introduced in \cite{bert}  and the model was fine-tuned on our dataset using Tensorflow Model Garden \cite{tensorflowmodelgarden2020}.
    \item Sentence-transformers \cite{sentence-bert} pre-trained MiniLM model \cite{minilm} was used to obtain 384-dimensional embeddings for documents and then SVM was used to classify the document embeddings.
    \item FinBERT model \cite{finbert} which is trained on financial text data for sentiment analysis was used to get 768-dimensional document embeddings and then SVM was used to classify the document embeddings.
\end{itemize}
Table \ref{tab:results} compares the final results of all the models developed in this study and our baselines. Link prediction task using DGI and Hinsage models have the highest f1-score among the others and has outperformed the baseline models.
\begin{table*}
    \centering
    \caption{Results}
    \label{tab:results}
    \begin{tabular}{ c c c c c }
    \toprule
    \textbf{Model} & \textbf{Task}  & \textbf{Precision} & \textbf{Recall} & \textbf{F1-score}\\
    \midrule
    DGI+HinSAGE & Link prediction & 0.9 & 0.89 & \textbf{0.89}\\
    HinSAGE & Link prediction & 0.89 & 0.88 & \textbf{0.89}\\
    Node2Vec\cite{node2vec}+Hinsage & Link prediction & 0.86 & 0.85 & 0.85\\
    Fine-tuned BERT & Document classification & 0.80 & 0.73 & 0.76\\
    SBERT embeddings+SVM & Document classification & 0.82 & 0.67 & 0.72\\
    DGI+HinSAGE & Node classification & 0.7 & 0.68 & 0.68\\
    HinSAGE & Node classification & 0.71 & 0.66 & 0.68\\
    finBERT+SVM & Document classification & 0.78 & 0.61 & 0.66\\
    \bottomrule
    \end{tabular}

\end{table*}
Link prediction methods, in general, outperform node classification by a large margin. By modeling the classification problem in the form of edge probability estimation, we are required to introduce a position node, and connect each tagged document to its corresponding position (label). We also connect some of the price patterns to their implications (long or short), which brings new kinds of information to this setting. Both modifications result in a more informative and richer graph structure that is essential for GNN models to learn better representations. We illustrate the output embedding of documents in 4 different settings in Figure \ref{fig:hinsage-nc}.

To examine the effectiveness of each group of nodes in the graph, experiments were done using different combinations of nodes. Table \ref{tab:experiments} summarizes the results for these experiments. The highest scoring result belongs to the link prediction task using HinSAGE model with word, document, topic, and position nodes. Using DGI and HinSAGE in the link prediction achieves almost the same result, however, it can be seen that DGI improves the performance of node classification in many cases. 
\def\checkmark{\tikz\fill[scale=0.4](0,.35) -- (.25,0) -- (1,.7) -- (.25,.15) -- cycle;}
\begin{table*}[t]
    \centering
    \caption{Experiments with different node settings}
    \label{tab:experiments}
    \begin{tabular}{ c c c c c c c c }
    \toprule
    \textbf{Model}& \textbf{Task}  & \textbf{Word} & \textbf{Document} & \textbf{Topic} & \textbf{Pattern} & \textbf{Position} &\textbf{F1-score}\\
    \midrule
    DGI + HinSAGE & Link Prediction & \checkmark & \checkmark & & & \checkmark& 0.885\\\hline
    DGI + HinSAGE & Link Prediction & \checkmark & \checkmark & \checkmark & & \checkmark & 0.89\\\hline
    DGI + HinSAGE & Link Prediction & \checkmark & \checkmark & & \checkmark & \checkmark & 0.885\\\hline
    DGI + HinSAGE & Link Prediction & \checkmark & \checkmark & \checkmark & \checkmark & \checkmark & 0.888\\\hline
    
    DGI + HinSAGE & Node Classification & \checkmark & \checkmark & & & & 0.68\\\hline
    DGI + HinSAGE & Node Classification & \checkmark & \checkmark & \checkmark & &  & 0.688\\\hline
    DGI + HinSAGE & Node Classification & \checkmark & \checkmark & & \checkmark &  & 0.68\\\hline
    DGI + HinSAGE & Node Classification & \checkmark & \checkmark & \checkmark & \checkmark &  & 0.512\\\hline

    HinSAGE & Link Prediction & \checkmark & \checkmark & & & \checkmark& 0.894\\\hline
    HinSAGE & Link Prediction & \checkmark & \checkmark & \checkmark & & \checkmark & 0.896\\\hline
    HinSAGE & Link Prediction & \checkmark & \checkmark & & \checkmark & \checkmark & 0.888\\\hline
    HinSAGE & Link Prediction & \checkmark & \checkmark & \checkmark & \checkmark & \checkmark & 0.888\\\hline
    
    HinSAGE & Node Classification & \checkmark & \checkmark & & & & 0.674\\\hline
    HinSAGE & Node Classification & \checkmark & \checkmark & \checkmark & &  & 0.652\\\hline
    HinSAGE & Node Classification & \checkmark & \checkmark & & \checkmark &  & 0.662\\\hline
    HinSAGE & Node Classification & \checkmark & \checkmark & \checkmark & \checkmark &  & 0.681\\
    
    \bottomrule
    \end{tabular}
\end{table*}

\section{Related works}\label{sec:5}

Graph-based approaches have recently gained attraction in natural language processing tasks. Graphs can express much more information in their structure than the bag-of-words approach. Therefore, representing text documents in  form of graphs could result in more accurate text classification. Text graphs can have different kinds of structures. Osman and Barukab \cite{reprreview} categorized graph-based representation approaches in natural language processing into 5 groups. For example, word co-occurrence graphs, document-word graphs, phrase as a graph, etc. Graph representation approaches can be divided into two groups: single graphs for a corpus that contain all documents \cite{yao2019graph, bertgcn} and individual graphs for each text in the document \cite{Shanavas2020, Jiang2010, texting, Zhao2021}. Graphs can also be heterogeneous or non-heterogeneous. Non-heterogeneous graphs contain only one type of nodes and node representations which is more common due to simplicity of the processing steps \cite{bertgcn, textgtl}. Heterogeneous graphs contain different types of nodes and representations \cite{Jiang2010, linmei-etal-2019-heterogeneous, hetegen}. A group of works in this area utilize graph-based approach for extracting embeddings that have captured similarities in the graph structures \cite{Shanavas2020, 5569866, dgrl}. However, graph structures can be used in the downstream tasks by turning the classification problem into node classification or link prediction problems \cite{yao2019graph}. Shanavas et al. \cite{Shanavas2020} used graph representation for each text document and defined a graph kernel for measuring similarity between graphs and then used SVM to classify documents. Jiang et al. \cite{Jiang2010} used a graph representation approach for documents with information of words like part of speech and their semantic features and then used graph mining approaches for feature vector extraction and classified the resulting vectors for each document.

In recent years, several deep learning methods have been proposed for graph-based text classification. For example, Yao et al. \cite{yao2019graph} leveraged Graph convolutional networks \cite{kipf} for text classification. They constructed a graph with relations between words and documents and turned the text classification problem into a node classification task. Zhang et al. \cite{texting} proposed TextING for inductive text classification task and trained a graph neural network that learns from detailed word-word relations from individual graphs for each document. Another method was proposed by Zhao and Huang et al. \cite{Zhao2021} for text classification using graphs for each document in the corpus. They used Bi-LSTMs to extract sequential features from graphs and then used graph convolutional networks for classification. Liu and You et al.  \cite{tensorgcn} constructed a text graph tensor and developed a learning method to harmonize information from multiple graphs of their graph tensor for text classification. In summary, this works builds on previous works in machine learning on graphs and explores the text classification task on a corpus represented as a heterogeneous graph. This study also applies text classification in finance for predicting investors' opinions which is new and different from previous sentiment analysis studies in this field.


\section{Conclusion}\label{sec:6}
We introduced a novel graph-based representation of stock market technical analysis reports for extracting financial insights and information. A heterogeneous graph with document, word, topic, price pattern, and position nodes was constructed from technical analysis documents such that each group of nodes had their specific initial node embeddings. Then, a graph neural network model was proposed to learn expressive representations for the node classification and edge prediction downstream tasks. Experimental results show that our model is capable of predicting labels accurately on our highly imbalanced dataset and outperforms the baseline methods. For the future direction, it is worthwhile to explore more types of nodes and relations for constructing more effective graph structures.

\begin{acks}
We would like to thank everyone at \textit{\textbf{Eveince}} for their valuable insights and support.
\end{acks}

\bibliographystyle{ACM-Reference-Format}
\bibliography{sample-base}

\end{document}